\documentclass[runningheads]{llncs}
\usepackage{xcolor}

\usepackage[T1]{fontenc}
\usepackage{graphicx}
\usepackage{booktabs} 
\usepackage{amsmath}
\usepackage{amssymb}
\usepackage{url}

\begin{document}
\title{Automatic Text Box Placement for Supporting Typographic Design}
%
%
\author{
Jun Muraoka\inst{1} 
\and Daichi Haraguchi\inst{2}\orcidID{0000-0002-3109-9053}  
\and Naoto Inoue\inst{2}\orcidID{0000-0002-1969-2006}
\and Wataru Shimoda\inst{2}\orcidID{0000-0001-6238-9697}
\and Kota Yamaguchi\inst{2}\orcidID{0000-0002-3597-2913}
\and Seiichi Uchida\inst{1}\orcidID{0000-0001-8592-7566}
}
%
\authorrunning{J. Muraoka et al.}
%
\institute{Kyushu University, Fukuoka, Japan\\
\email{jun.muraoka@human.ait.kyushu-u.ac.jp}
\email{uchida@ait.kyushu-u.ac.jp}
\and
CyberAgent, Tokyo, Japan\\
\email{\url{{haraguchi_daichi_xa, inoue_naoto, wataru_shimoda, yamaguchi_kota}@cyberagent.co.jp}}}
\maketitle              
\begin{abstract}
In layout design for advertisements and web pages, balancing visual appeal and communication efficiency is crucial. This study examines automated text box placement in incomplete layouts, comparing a standard Transformer-based method, a small Vision and Language Model (Phi3.5-vision), a large pretrained VLM (Gemini), and an extended Transformer that processes multiple images. Evaluations on the Crello dataset show the standard Transformer-based models generally outperform VLM-based approaches, particularly when incorporating richer appearance information. However, all methods face challenges with very small text or densely populated layouts. These findings highlight the benefits of task-specific architectures and suggest avenues for further improvement in automated layout design.

\end{abstract}
%
%
\section{Introduction\label{sec:intro}}

\begin{figure}[t]
\includegraphics[width=\textwidth]{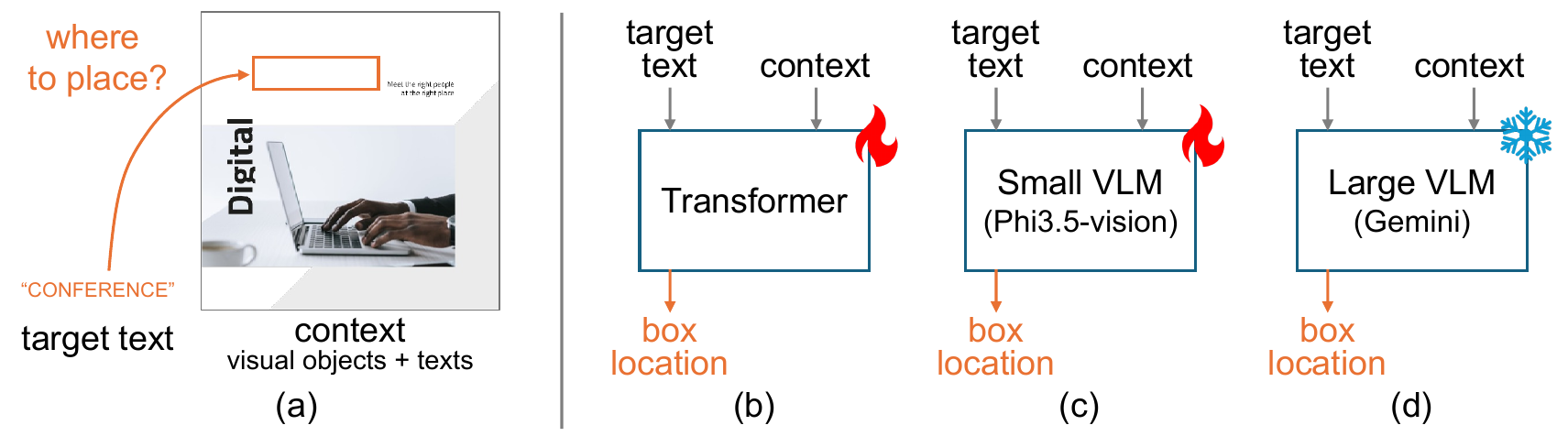}
\caption{(a)~Overview of the text box placement task. (b)-(d)~Three machine learning-based methods for solving the task.} \label{fig:models}
\end{figure}

Layout design plays a crucial role in various forms of media, such as advertisements, posters, and web pages, by organizing information in a visually clear and comprehensible manner. In our digital society, where information overload is a constant challenge, effective layout design enhances both visual appeal and information delivery. However, achieving an optimal balance in the placement, size, and color harmony of design elements is a complex and time-intensive process, requiring both aesthetic sensibility and iterative refinements. Ineffective arrangements can hinder communication, highlighting the need for technologies that streamline the layout design process.
\par

Modern design formats often employ a layered structure, allowing elements to be independently adjusted. This flexibility enables seamless modifications and refinements, making it particularly valuable for multimodal layouts that integrate text and images. While images provide visual impact, text conveys detailed information, and their harmonious integration is crucial for clear communication. Automated layout generation has gained attention in research due to its potential to enhance efficiency and adaptability in design workflows.
\par

The purpose of this study is to understand how recent methodologies can support typographic design through a simple but {\em very essential layout task}: that is, the optimal placement for incorporating a new text box into partially completed multimodal layouts. Fig.~\ref{fig:models}(a) illustrates this estimation task. By leveraging the layered structure of modern layouts, elements can be dynamically adjusted to maintain visual coherence and structural integrity. Addressing this task is expected to improve both the efficiency and quality of the layout design process.
\par

As possible methodologies, we implement and evaluate three methods for estimating optimal text box placements,  as shown in Figs.~\ref{fig:models}(b), (c), and (d): 
\begin{itemize}
    \item A standard Transformer-based method~(b), which directly trains the text placement task as a regression problem, considers relationships among a set of various elements, such as the whole image, text elements, and image elements.
    \item A small Vision and Language Model (VLM)-based method~(c), which utilizes Phi3.5-vision~\cite{abdin2024phi}, a lightweight VLM. This method relies on a single global image and estimates placements with a reduced set of input features compared to the standard Transformer-based method. Additionally, we examine whether placement estimation can be improved by controlling prompt inputs.
    \item A large VLM-based method~(d), which leverages the large-scale pretrained model, Gemini~\cite{team2023gemini}, without additional training. For this method, we investigate its ability to perform estimation under the constraint of no additional training. 
\end{itemize}
\par

By comparing these three methods as fairly as possible, this study helps to understand which methodology is more promising for the optimal text box placement task. More importantly, this study helps to identify scenarios that are inherently easy or difficult for all three methods. A typical case that is easy for all methods suggests the presence of universal constraints that limit placement flexibility, leading to a relatively deterministic solution. Conversely, cases that are difficult for all methods indicate situations where placement freedom is extremely high, making the optimal solution ambiguous or even non-unique. Understanding these inherent constraints and degrees of freedom in layout composition contributes to a deeper theoretical insight into automated text box placement.\par
This study can also be viewed as developing new application directions, such as appending additional text to AI-generated typographic layouts, supporting interactive text-manipulation processes, verifying the
placement of individual text boxes, etc. Note that this study elucidates the effectiveness of a task-tailored training method, a fine-tuned VLM, and a fully-pretrained VLM in capturing spatial relationships and integrating new elements into existing designs.

\section{Related Work\label{sec:related}}

\subsection{Layout generation using basic machine learning methods}


Layout generation has been extensively studied in recent years. Content-agnostic approaches, which do not consider specific content, have been explored using Generative Adversarial Networks (GANs)~\cite{li2019layoutgan}, Variational Autoencoders (VAEs)~\cite{jyothi2019layoutvae}, and diffusion models~\cite{inoue2023layoutdm}. However, these methods face limitations in practical scenarios as they do not incorporate specific content like text or images into their design considerations.
\par
More recently, content-aware layout generation, which accounts for background images or text, has gained attention. For instance, FlexDM~\cite{inoue2023document} is a Transformer-based model that predicts masked elements and their attributes. Similarly, RADM~\cite{li2023relation} utilizes diffusion models to generate typed bounding boxes over background images, and RALF~\cite{horita2024retrieval} enhances a Transformer framework with retrieval-augmented layout information.
\par

\subsection{Layout generation using fine-tuned VLMs for the task}


The use of fine-tuned VLMs for layout generation has garnered attention due to their scalability. COLE~\cite{jia2023cole} introduced a pipeline to generate layouts from natural language instructions using a VLM to predict attributes like text placement. PosterLLaVA~\cite{yang2024posterllava} also utilizes a multimodal LLM to position elements based on user prompts, while PosterLlama~\cite{seol2024posterllama} employs a novel depth-based augmentation strategy to enhance robustness in data-scarce scenarios.
\par
In this study, we leverage Phi3.5-vision~\cite{abdin2024phi}, a lightweight VLM. Its smaller size and computational efficiency make it suitable for fine-tuning on our single text box placement task, while retaining sufficient expressive power to learn spatial relationships.
\par

\subsection{Layout generation using large-scale VLMs without training} 

Layout generation methods utilizing large-scale VLMs without additional training have also been proposed. While large-scale models, such as GPTs~\cite{achiam2023gpt}, pose challenges in terms of high computational costs and limited accessibility for fine-tuning, they have already been trained on vast and diverse knowledge sources. As a result, they may have implicitly acquired substantial knowledge related to layout design. Particularly, large-scale VLMs like Gemini~\cite{team2023gemini} can process both text and images, making them highly relevant for layout tasks that require an understanding of two-dimensional spatial structures. This potential for high performance without task-specific tuning makes them an important research direction.
\par
LayoutPrompter~\cite{lin2024layoutprompter} demonstrated the feasibility of performing various tasks solely through prompt engineering with pretrained LLMs. It achieved high levels of accuracy by incorporating mechanisms such as dynamic few-shot selection through retrieval and a Ranker module to evaluate multiple candidate outputs. Similarly, LayoutGPT~\cite{feng2024layoutgpt} introduced a framework that generates not only 2D layouts but also 3D configurations flexibly, based on user instructions.

Let us conclude this review section with the reason why we need to conduct our research on the optimal single text box placement, although we can find several existing methods that estimate bounding boxes for all elements or perform multitask inferences for various attributes. The existing methods tend to evaluate overall accuracies, leaving the fundamental constraints and degrees of freedom in text placement underexplored. In contrast, we aim to clarify these constraints and better understand the inherent complexity of the task, by focusing on a single text box. Additionally, although various layout generation techniques exist, no comprehensive study has compared SOTA VLMs under a unified experimental setting. This work fills that gap by evaluating a Transformer-based model, a fine-tuned VLM, and a large-scale pretrained VLM, providing insights into their relative strengths, limitations, and applicability to automated layout design.

\section{Three Machine Learning-Based Methods for Text Box Placement}



This section describes the three methods selected for estimating the optimal placement of text boxes in layouts. As discussed in Section~\ref{sec:related}, layout generation has been approached through three major directions: (1) basic machine learning models tailored for layout tasks, (2) fine-tuned VLMs, and (3) large-scale pretrained VLMs used without additional training. To comprehensively evaluate their applicability to the single text box placement task, we select a representative method from each category: 
\begin{itemize}
    \item A standard Transformer-based method, which directly models spatial relationships between multimodal elements through task-specific training (Section~\ref{sec:method1}).
    \item A fine-tuned VLM-based method, which adapts Phi3.5-vision~\cite{abdin2024phi} to process layout information as structured language input (Section~\ref{sec:method2}).
    \item A large-scale pretrained VLM-based method, which leverages Gemini~\cite{team2023gemini} without additional training to assess its zero-shot capabilities (Section~\ref{sec:method3}).
\end{itemize}
By comparing these methods, we aim to assess not only their relative performance but also the fundamental challenges and constraints inherent in the single text box placement task.
\par

All these methods estimate the optimal placement of a new text box within an existing layout. As illustrated in Fig.~\ref{fig:model-TF}(a), the layout consists of $N$ elements, where one element corresponds to the target text to be placed, and the remaining $N-1$ elements represent existing layout components as context. Note that one of $N-1$ contextual elements is the whole layout image (given by rendering all elements except for the target text).
\par

\subsection{Standard Transformer-based method~\label{sec:method1}}
\begin{figure}[t]
\includegraphics[width=\textwidth]{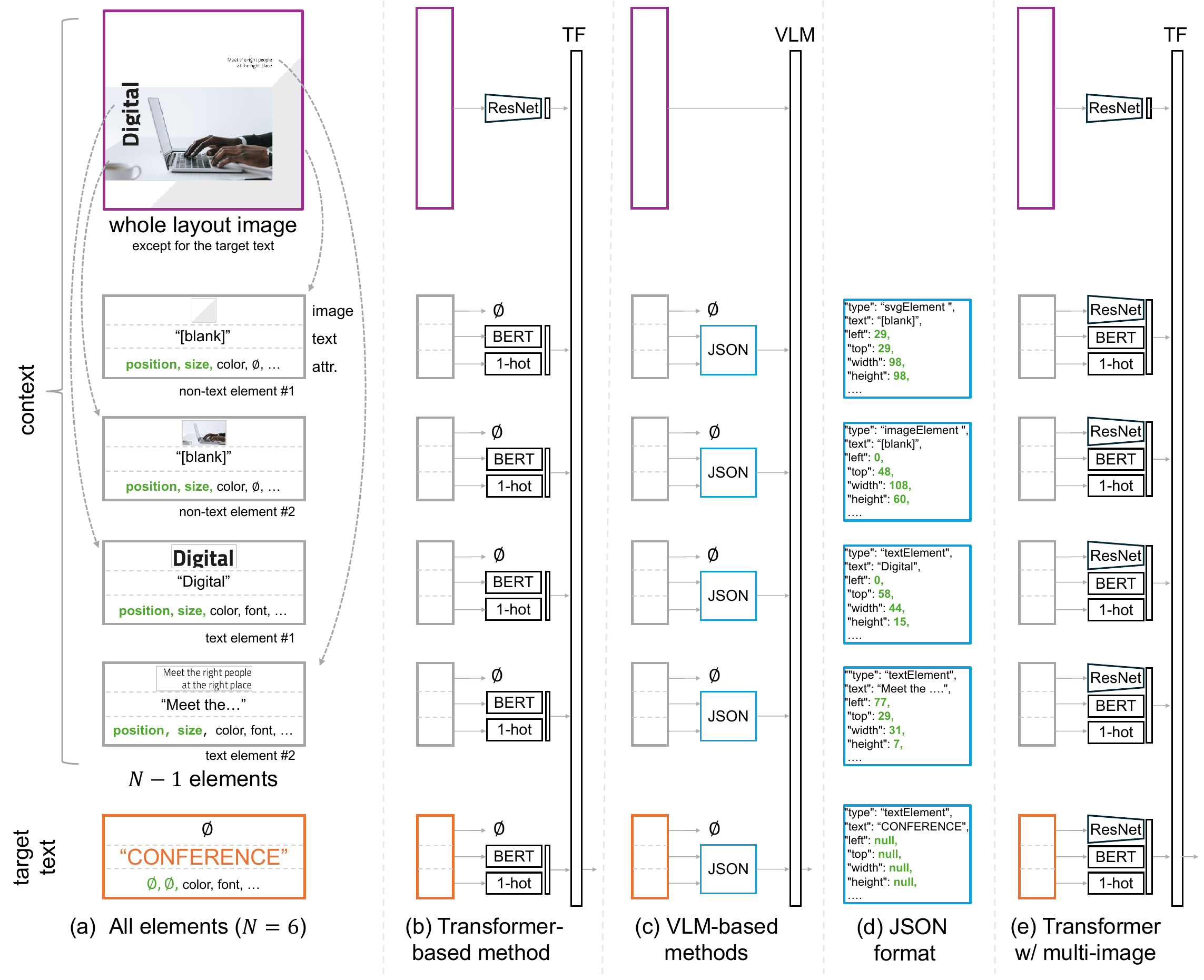}
\caption{(a)~Details of $N$ input elements. (b) and (c)~Models for the optimal text box placement. (d)~The details of the JSON format for VLM input. (e)~Another version of (b) for utilizing a bitmap-based representation of individual input elements.} \label{fig:model-TF}
\end{figure}


The standard Transformer-based method estimates the placement of a target text box by processing the layout information of all $N$ elements, as shown in Fig.~\ref{fig:model-TF}(b). The model takes as input a set of feature representations for these elements and outputs the coordinates of the text box to be placed.  The coordinates are given as a four-dimensional vector whose elements correspond to left, top, width, and height. Note that these values are normalized to the range $[0,1]$, ensuring compatibility with various document sizes. 
\par

This method offers several advantages. First, the self-attention mechanism enables the model to learn spatial relationships between $N$ elements, making it well-suited for layout tasks where interactions between text and surrounding objects are critical. Second, the Transformer architecture can handle an arbitrary number of input elements, allowing for greater flexibility in varying layout compositions. Third, by incorporating appropriate encoders, the model can seamlessly process diverse input types, including images, raw text, and categorical attributes represented as one-hot vectors, making it adaptable to multimodal layout representations.
\par
As illustrated in Fig.~\ref{fig:model-TF}(b), the Transformer encoder processes a set of $N$ layout elements consisting of three types of inputs. First, the {\em target text} is represented by its rendered image, text content, character count, line count, angle, color, and font, but without location attributes, as these are the targets of our estimation. Second, the remaining $N-1$ elements in the layout include similar attributes --- rendered images, text content, position, size, type, angle, color, and font --- where non-text elements are assigned empty strings or zeros for text-specific attributes, and categorical data such as font and type are encoded as one-hot vectors. Third, the model also takes as input the whole layout image with the target text removed.
\par
To process these inputs, image and text data are first encoded using a ResNet50-based image encoder and a BERT-based text encoder, respectively. The embeddings obtained from these encoders, along with other layout attributes, are compressed through linear transformations and concatenated for each element before being fed into the Transformer encoder. 
\par

\subsection{Small VLM-based method\label{sec:method2}}



Fig.~\ref{fig:model-TF}(c) shows the structure of the VLM-based method. Here, we use a ``small'' (light-weight) VLM to handle images and texts jointly. We employ Phi3.5-vision~\cite{abdin2024phi}, which has already pretrained knowledge. Its lightweight nature facilitates fine-tuning for estimating the optimal text box placements even with limited training data. Its inputs (texts and one-hot attributes) are the same as in the standard Transformer-based method; however, instead of encoding them by BERT and linear transformations, 
the inputs are converted in JSON format, as illustrated in Fig.~\ref{fig:model-TF}~(d), and fed to the Phi3.5-vision as a structured prompt. While the standard Transformer-based method predicts continuous values for text box placement, VLMs predict text box placement in a linguistic format, i.e., a sequence of tokens.\par

The JSON structure describes all the contextual elements and the target text using the fields: {\tt type}, {\tt text}, {\tt left}, {\tt top}, {\tt width}, and {\tt height}. The field {\tt type} describes the type of element. In this paper, we use one type for texts and four types for images.\footnote{These five types are defined in the Crello dataset~\cite{yamaguchi2021canvasvae} as follows: {\tt textElement} for texts, {\tt imageElement} for bitmap images,  {\tt maskElement} for bitmap images with cropping by a mask, {\tt svgElement} for shapes or textures in the SVG (i.e., vector) format, and {\tt coloredBackground} for a constant-color background.} The field {\tt text} describes the text sequence of each element; for non-text elements, this field is given as the special placeholder {\tt \{blank\}}. The remaining {\tt left}, {\tt top}, {\tt width}, and {\tt height} specify the size and location of the bounding box for each element. Note that these four fields are set to {\tt null} for the target text. The JSON descriptions of individual elements are sorted by their position and fed to the VLM. Finally, the VLM model generates output in JSON format, containing the estimated left, top, width, and height.

\subsection{Large VLM-based method~\label{sec:method3}}
Fig.~\ref{fig:model-TF}(c) also shows the method with a large VLM; the differences from the small VLM-based method are their model size. The large VLM-based method uses Gemini~\cite{team2023gemini}, which is much larger than Phi3.5-vision. Although Gemini is not trainable with our rather small dataset, it is already pretrained with vast knowledge. Therefore, even without any additional training, we can expect that Gemini already has sufficient knowledge about typography and layout patterns. We, consequently, evaluate Gemini with the same inputs and output as the small VLM-based method, illustrated in Figs.~\ref{fig:model-TF}(c) and (d). 

\section{Experimental Results}
\subsection{Dataset\label{sec:dataset}}
Professionally designed and layer-segmented typography datasets are scarce, and therefore, we used
the Crello dataset~\cite{yamaguchi2021canvasvae}, the best resource currently available.
This dataset consists of professionally designed layouts, including social media posts, printed posters, and banner advertisements, all represented in a layered structure. Each design was created by typographic experts and is stored as a multi-layered file, similar to those used in Illustrator or other vector-based design software.
\par

The dataset originally contains 19,095, 1,951, and 2,375 layouts for the training, validation, and test sets, respectively. Since our study focuses on estimating the position and size of a single text string, we excluded designs that do not contain textual elements. After this exclusion, the dataset includes 18,937, 1,933, and 2,314 layouts for training, validation, and testing, respectively. For the quantitative evaluation, we further divide the test set into 173 layouts with a single text element and 2,141 layouts with multiple text elements, following previous works~\cite{inoue2023document,li2021harmonious}.

\subsection{Training details}
\subsubsection{Standard Transformer-based method}
We used only the encoder of the Transformer model. The Transformer encoder consists of six layers. Each layer consists of eight attention heads, each with a dimension of $256$, and a feed-forward network with a dimension of $512$.
The model was trained using the AdamW optimizer, with a learning rate of $0.0001$ and a batch size of 64. After $150$ epochs of training, we use the model with the lowest validation loss.\par

We use Complete IoU (CIoU) loss~\cite{zheng2020distance} to optimize the model. CIoU is an extended version of the standard IoU (Intersection over Union) that incorporates three key factors: the standard IoU between the predicted and ground-truth boxes, the distance between their centroids, and the consistency of aspect ratio. By accounting for centroid distance, CIoU mitigates the issue where standard IoU drops to zero when small boxes are slightly misaligned and do not overlap at all. Additionally, CIoU considers aspect ratio to ensure that the predicted bounding box maintains a shape similar to the ground-truth box. This is particularly important in text placement tasks, where bounding box proportions influence visual harmony and spatial alignment within the layout.

\subsubsection{Small VLM-based method (Phi3.5-vision)}
To fine-tune Phi3.5-vision \cite{abdin2024phi}, we employed LoRA (Low-Rank Adaptation) \cite{hu2021lora}, a common technique for adapting LLMs and VLMs, with a rank of $128$ and an alpha of $256$. We trained the model using cross-entropy loss and optimized it with AdamW. The learning rate was set to $0.00005$ and the batch size to $1$. Training was conducted over five epochs, and we selected the model achieving the lowest validation loss for evaluation.

\subsubsection{Large VLM-based method (Gemini)}
For the large-scale pretrained model, we employed Gemini 2.0 Flash (model identifier: gemini-2.0-flash-exp)\cite{team2023gemini} in a zero-shot manner, i.e., without any additional training. As already explained,  we structured the model’s inputs and outputs in JSON format, facilitating more reliable parsing and evaluation.

\subsection{Evaluation metrics}
\begin{table}[t]
\centering
\caption{Quantitative evaluation of text box placement accuracy.
\textbf{Bold} and \underline{underline} indicate the best and second-best results, respectively. Following prior works~\cite{inoue2023document,li2021harmonious}, we split the test data into layouts containing a single text element (``Single text'') or multiple text elements (``Multiple text'').}
{\tabcolsep = 7pt
\begin{tabular}{@{}llcccccc@{}}
\toprule
& Input & \multicolumn{2}{c}{Single text (173)} & \multicolumn{2}{c}{{Multiple text (2,141)}} & \multicolumn{2}{c}{{All (2,314)}}\\
& image  & {IoU$\uparrow$} & {BDE$\downarrow$} & {IoU$\uparrow$} & {BDE$\downarrow$} & {IoU$\uparrow$} & {BDE$\downarrow$}\\
\midrule

Transformer  & Single &\textbf{0.269} & \textbf{0.131} &\textbf{0.261} & \textbf{0.121} & \textbf{0.262} & \textbf{0.122}\\
Phi3.5-vision & Single & 0.204 & 0.177 & 0.200 & 0.126 & 0.200 & 0.130\\
Gemini &  Single& 0.140 & 0.178& 0.133 & 0.181 & 0.134 & 0.181\\ \hline\hline
 FlexDM\cite{inoue2023document}$\dagger$ & Multi$\star$ & 0.357 & 0.098 & 0.110 & 0.141 \\  
Transformer & Multi$\star$ & \textbf{0.365} & \textbf{0.096} & \textbf{0.295} & \textbf{0.104} & \textbf{0.300} & \textbf{0.104} \\
\bottomrule
\multicolumn{6}{l}{$\dagger$: Values taken from results in \cite{inoue2023document}.\ \  $\star$: Discussed later in Section~\ref{sec:advantage_multi}}
\end{tabular}\label{tab:comparison_table}}
\end{table}

We employed the standard IoU and Boundary Displacement Error (BDE) \cite{li2021harmonious} to evaluate how closely the predicted results align with the actual layout. BDE is similar to the centroid distance in CIoU. Specifically, BDE measures how far the edges of the predicted bounding box deviate from those of the ground-truth box. Specifically, it looks at the differences in position between the left, right, top, and bottom edges of both boxes and then averages those differences. A lower BDE value indicates closer alignment between the predicted and true boundaries, whereas a higher value suggests a larger mismatch.\par

Note that the VLM-based methods occasionally generate outputs that do not conform to the expected format, rendering them unsuitable for evaluation. Specifically, Phi3.5-vision produced 7 invalid outputs (out of 2,313), while Gemini produced 18. We excluded these cases from our analyses.

\subsection{Quantitative evaluation}
\subsubsection{Quantitative evaluation of overall performance}\label{sec:overall_quantitative}

The results obtained using the aforementioned metrics are summarized in Table~\ref{tab:comparison_table}.  As noted in Section~\ref{sec:dataset}, following prior work, we split the test set into ``Single text'' and ``Multiple text'' layouts. In the Single text subset, no other text elements are present, so the model is free from overlap concerns but also lacks the guidance that existing text might provide. Conversely, multiple text subsets provide more hints about where to place new text because other text elements are already laid out, but they add complexity to avoiding unnecessary overlaps between texts and maintaining a consistent layout.
\par
 Table~\ref{tab:comparison_table} shows that the standard Transformer-based model consistently achieves the highest IoU and BDE scores across both Single-text and Multiple-text tasks. We attribute this superiority to two main factors. First, the Transformer is trained end-to-end specifically for text box placement, directly regressing continuous bounding-box coordinates and learning the spatial constraints needed for precise layout. Second, while Phi3.5-vision undergoes fine-tuning for this task and therefore outperforms Gemini, which is used in a purely zero-shot manner, both are general-purpose VLMs originally trained on broader objectives. Their token-based coordinate generation may also introduce discretization issues, causing slight format or numerical mismatches. Consequently, even the fine-tuned VLM struggles to capture the finer-grained relationships required for high-accuracy text positioning, highlighting the advantages of a specialized approach.
 \par
 
\begin{figure}[t]
\centering
\includegraphics[width=0.9\textwidth]{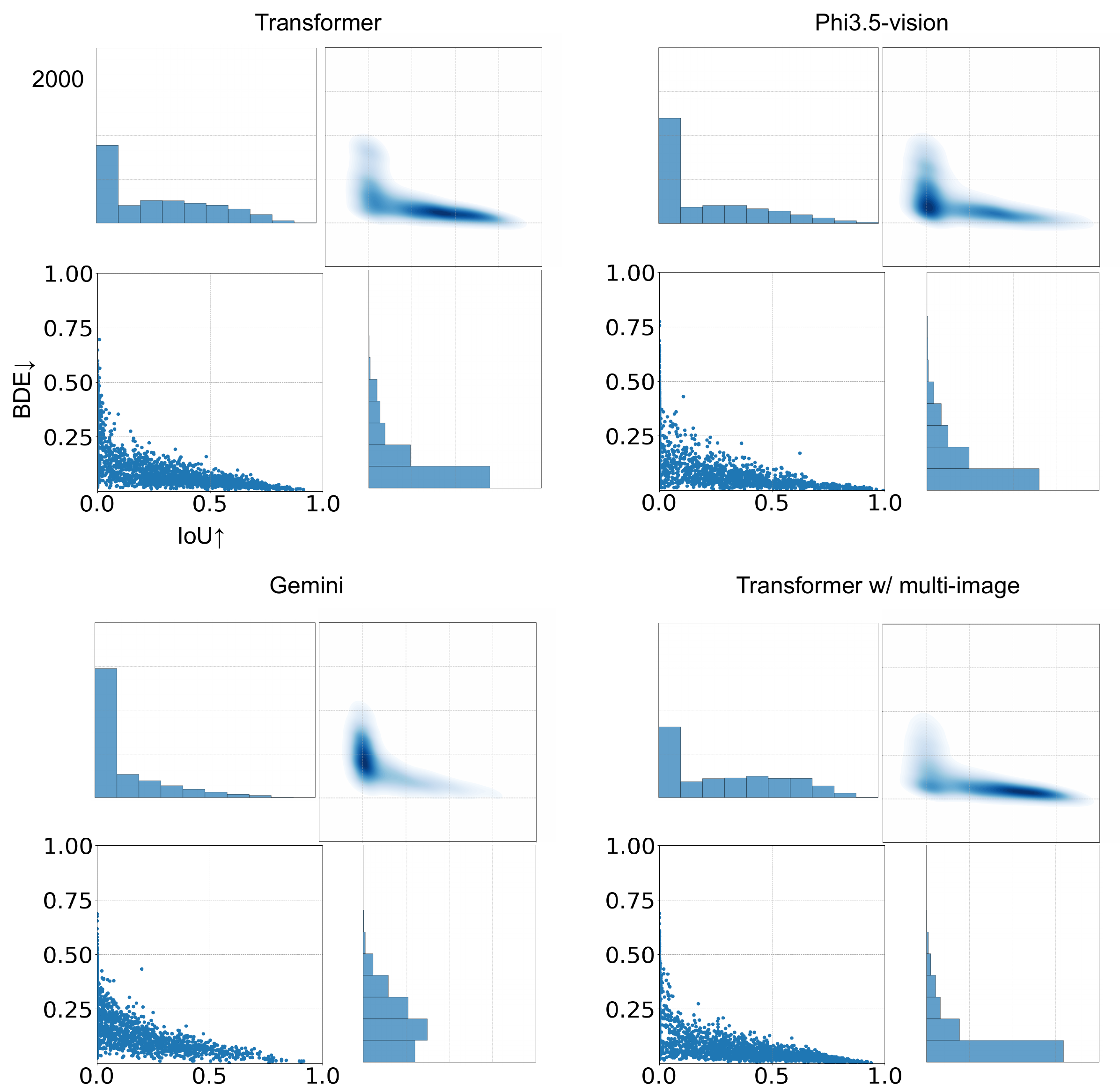}\\[-3 mm]
\caption{Distributions of IoU and BDE in all the test set (single text + multiple text). We present IoU and BDE histograms, an IoU-BDE scatter plot, and a corresponding heatmap for each method.}
\label{fig:yyplot}
\end{figure}



Fig.~\ref{fig:yyplot} shows the histograms of IoU and BDE for each method, as well as an IoU-BDE scatter plot and its heatmap representation. The IoU histograms reveal that near-zero IoU values occur relatively frequently across all methods. However, this does not mean the placement task is intractable. In practice, IoU should be considered alongside BDE: a large number of cases with both near-zero IoU and small BDE often indicate small text boxes that barely fail to overlap due to minor misalignments. Indeed, cases with large BDE (and thus an IoU of zero) remain relatively rare for all methods, suggesting that text placement is still reasonably accurate in most cases.
\par

When comparing methods, the Transformer and Phi3.5-vision methods exhibit pronounced peaks near zero in their BDE histograms, indicating that most predictions have only minor displacements. Meanwhile, Gemini’s BDE distribution does not peak around zero, reflecting the model’s inability to outperform those specialized or fine-tuned for text placement. Additionally, the frequency of high-precision placements (i.e., high IoU values) decreases monotonically for the two VLM-based methods. (In short, better IoU is less frequent.) By contrast, the Transformer’s IoU histogram shows a secondary peak around IoU$\sim$ 0.3, deviating from the strictly monotonic trend.

\begin{figure}[t]
\centering\centering
\includegraphics[width=\textwidth]{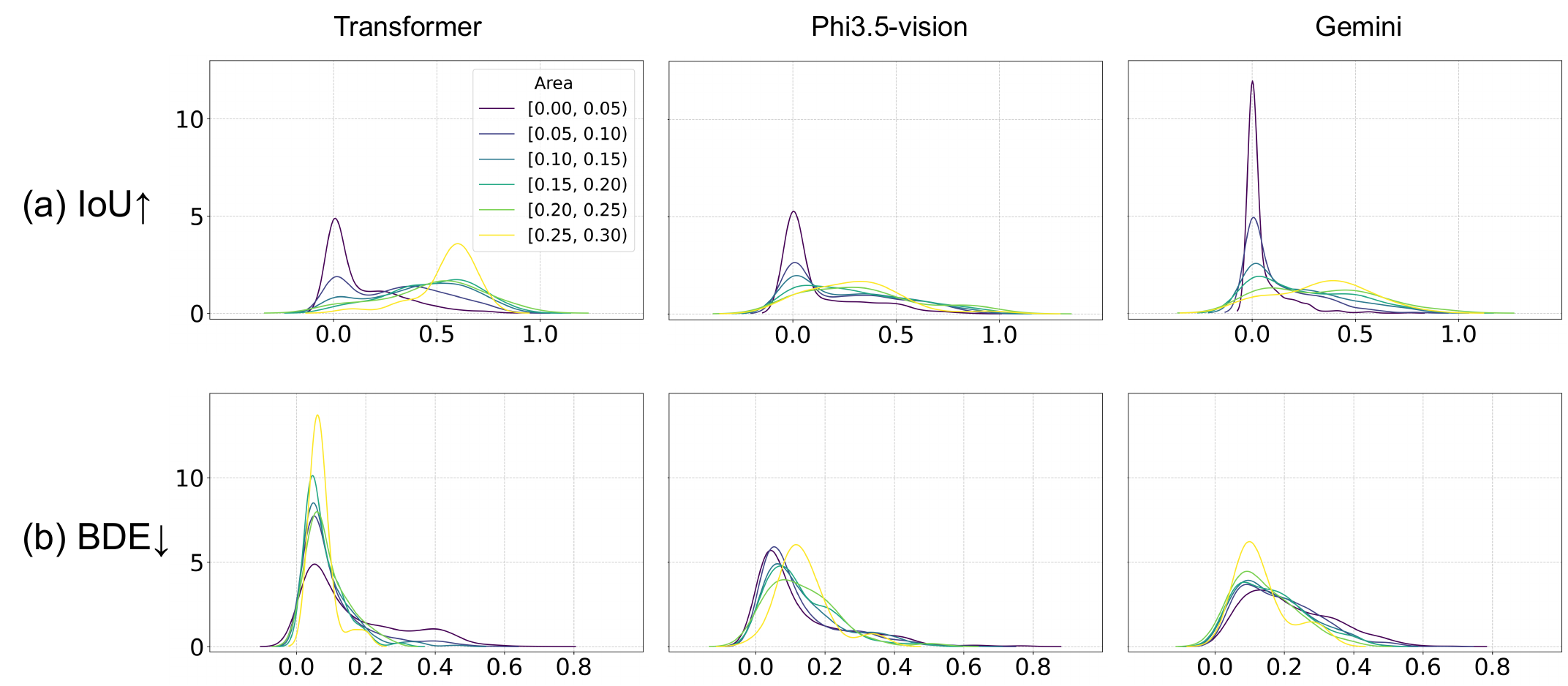}\\[-3mm]
\caption{Effect of the target text area on placement accuracy.} \label{fig:kde_area}
\bigskip
\includegraphics[width=1.0\textwidth]{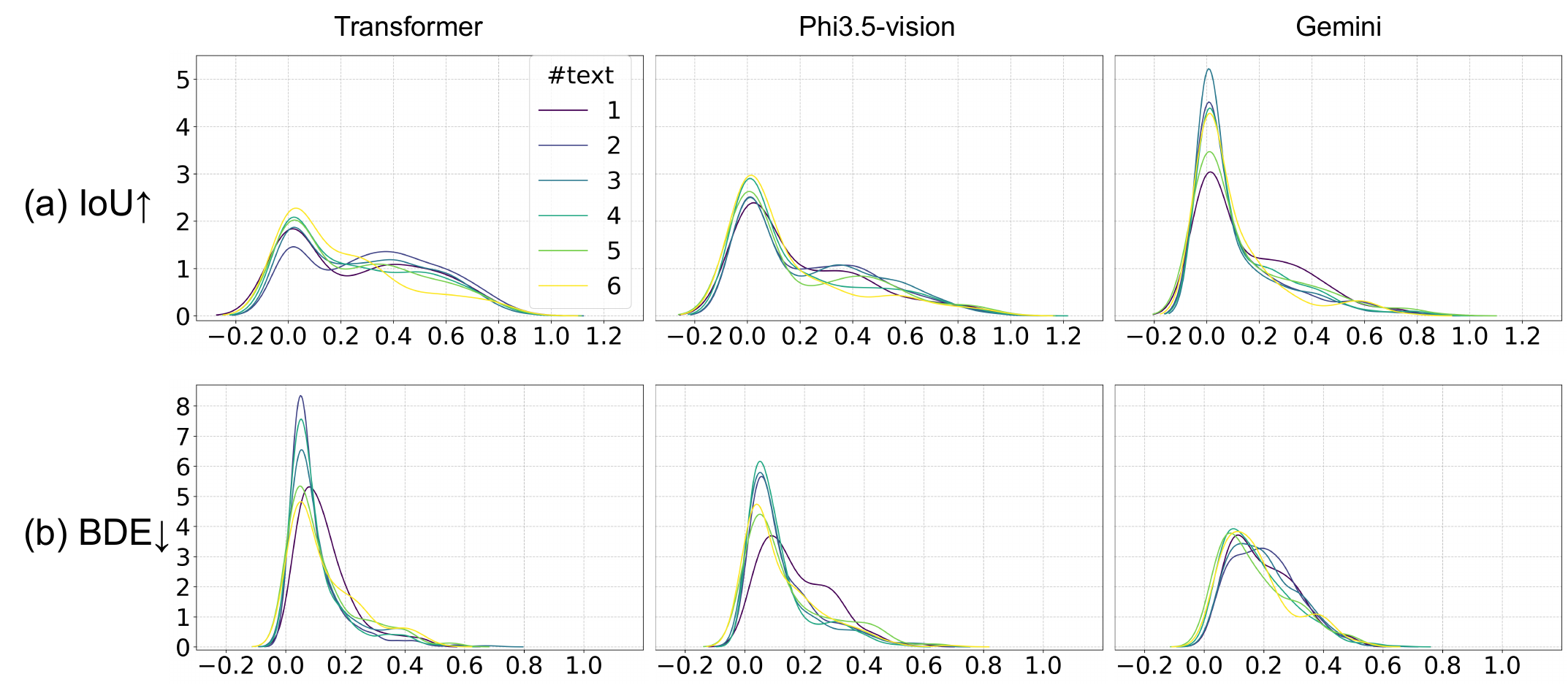}\\[-3mm]
\caption{Effect of the number of text elements on placement accuracy.\label{fig:kde}}
\end{figure}

\subsubsection{Effect of the target text area on placement accuracy}

Fig.~\ref{fig:kde_area} shows how the target text area affects placement accuracy for each method, visualized via Kernel Density Estimation (KDE). The purple curves represent the smallest text areas, transitioning to yellow for progressively larger areas. Although IoU and BDE are each bounded within $[0,1]$, KDE smoothing can produce density estimates that appear outside this interval.
\par
Across all methods, smaller text areas lead to poorer placement accuracy in terms of both IoU and BDE. This is partly due to the nature of IoU, which inherently penalizes small bounding boxes, resulting in near-zero IoU values especially for Gemini. By contrast, larger text areas tend to improve both IoU and BDE; however, an exception emerges with Phi3.5-vision, which shows a tendency toward higher (i.e., worse) BDE as the target text area increases.

\subsubsection{Effect of the number of text elements on placement accuracy}
Fig.~\ref{fig:kde} shows KDE plots illustrating how the number of text elements in a layout influences placement accuracy for each method. The purple curve (\#text = 1), represents a single-text layout with only the target text present. Cases where \#text 
$\geq$2 indicate one or more additional text elements besides the target text.
\par

Generally, as the number of text elements increases, IoU tends to decline because each text box becomes smaller in general. However, this decrease is not always monotonic. IoU starts off low at \#text = 1 for the Transformer model, improves noticeably at \#text = 2, and then exhibits more near-zero IoU as additional texts are added. A similar pattern is observed in the BDE metric, suggesting that having at least one other text element may help guide the placement of the target text.
\par

In contrast, Phi3.5-vision's IoU remains relatively low and shows little change across different text counts. However, its BDE behavior mirrors that of the Transformer: the highest accuracy occurs when there are a few additional text elements (e.g., $2\sim 4$), indicating that a moderate number of contextual texts can benefit the placement process.

\begin{figure}[t]
\includegraphics[width=\textwidth]{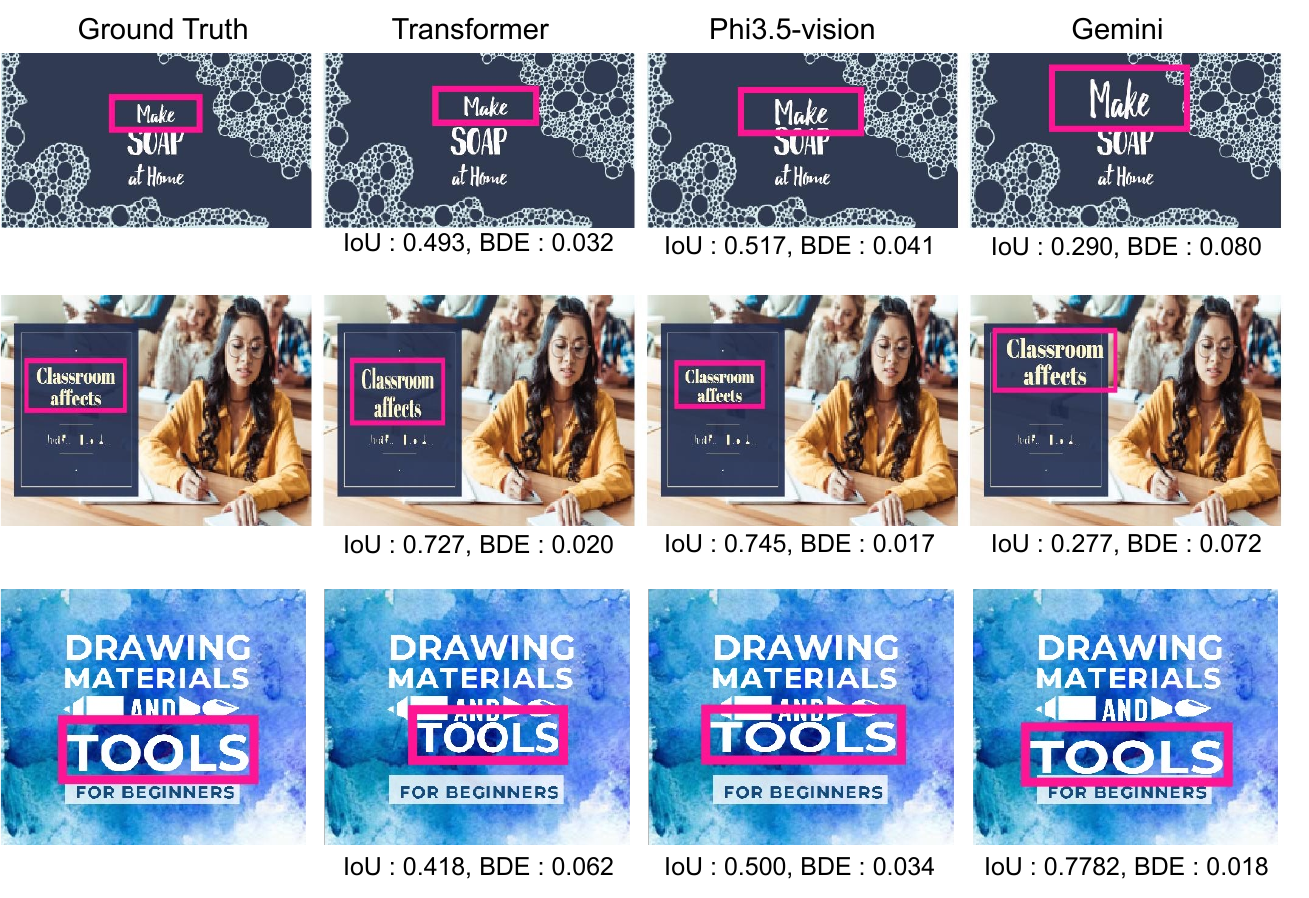}\\[-8mm]
\caption{Representative examples where all three methods placed the text box successfully. \label{fig:result_images}}
\end{figure}
\begin{figure}[t]
\includegraphics[width=\textwidth]{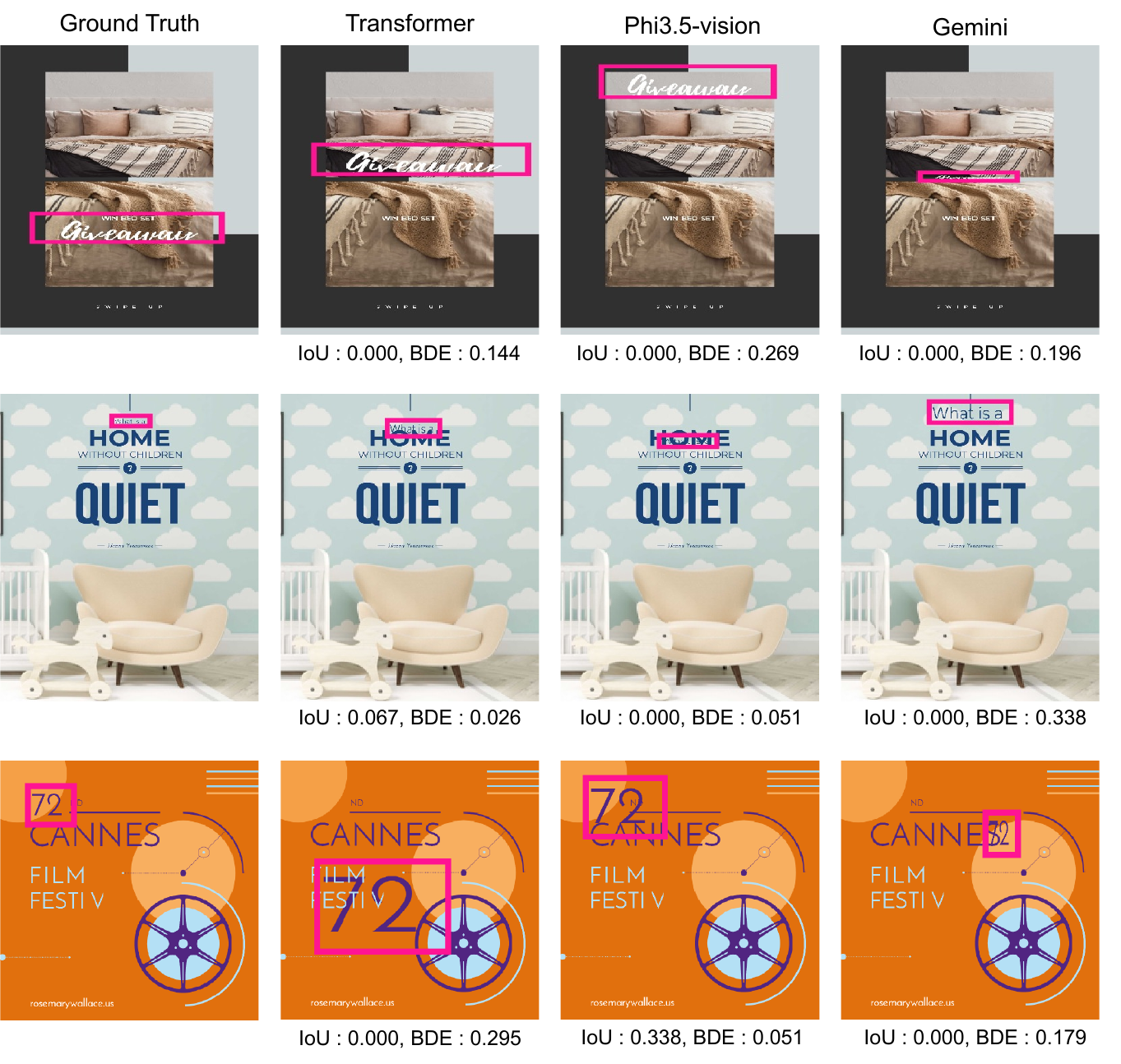}\\[-5mm]
\caption{Representative failure cases in all three methods. (Top) layouts where the text must be placed on a visually prominent area of the image, (middle) small text boxes requiring precise spacing, and (bottom) dense layouts that lead to unintended overlaps.} \label{fig:failure_images}
\end{figure}

\subsection{Qualitative evaluation}


\subsubsection{Successful cases}
Fig.~\ref{fig:result_images} shows examples in which all three methods achieved consistently high-quality placements. In the top row ($N=3$), the text ``Make'' seamlessly aligns with the existing text ``SOAP at Home.'' In the second row ($N=2$), the target text ``classroom\textbackslash n affects'' (where a newline token is inserted between the two words) is accurately placed over a sufficiently large background image with clearly defined design elements. A satisfactory placement was frequently observed in such large background scenarios.
This result also demonstrates the methods’ ability to handle multi-line inputs effectively. Finally, the bottom row ($N=5$) demonstrates that all three methods performed precise and contextually appropriate placements even when the target text must fit between existing text elements. Notably, even without task-specific fine-tuning, Gemini placed the bounding boxes well, showing a notable level of performance.
\par

\subsubsection{Failure cases}
Fig.~\ref{fig:failure_images} shows representative failure cases observed in all three methods, focusing on three typical scenarios. The top row illustrates layouts where the text must be placed on a non-uniform background, yet models are typically trained to place text on flatter regions for better readability; limited data on such complex cases makes it difficult to learn appropriate placement. The middle row depicts small text boxes, which demand near-pixel-perfect alignment in tight whitespace. The bottom row presents dense layouts with limited free space, often forcing unintended overlaps. Even with dedicated or fine-tuned models, these edge conditions pose substantial challenges.

\begin{figure}[t]
\centering
\includegraphics[width=1.0\textwidth]{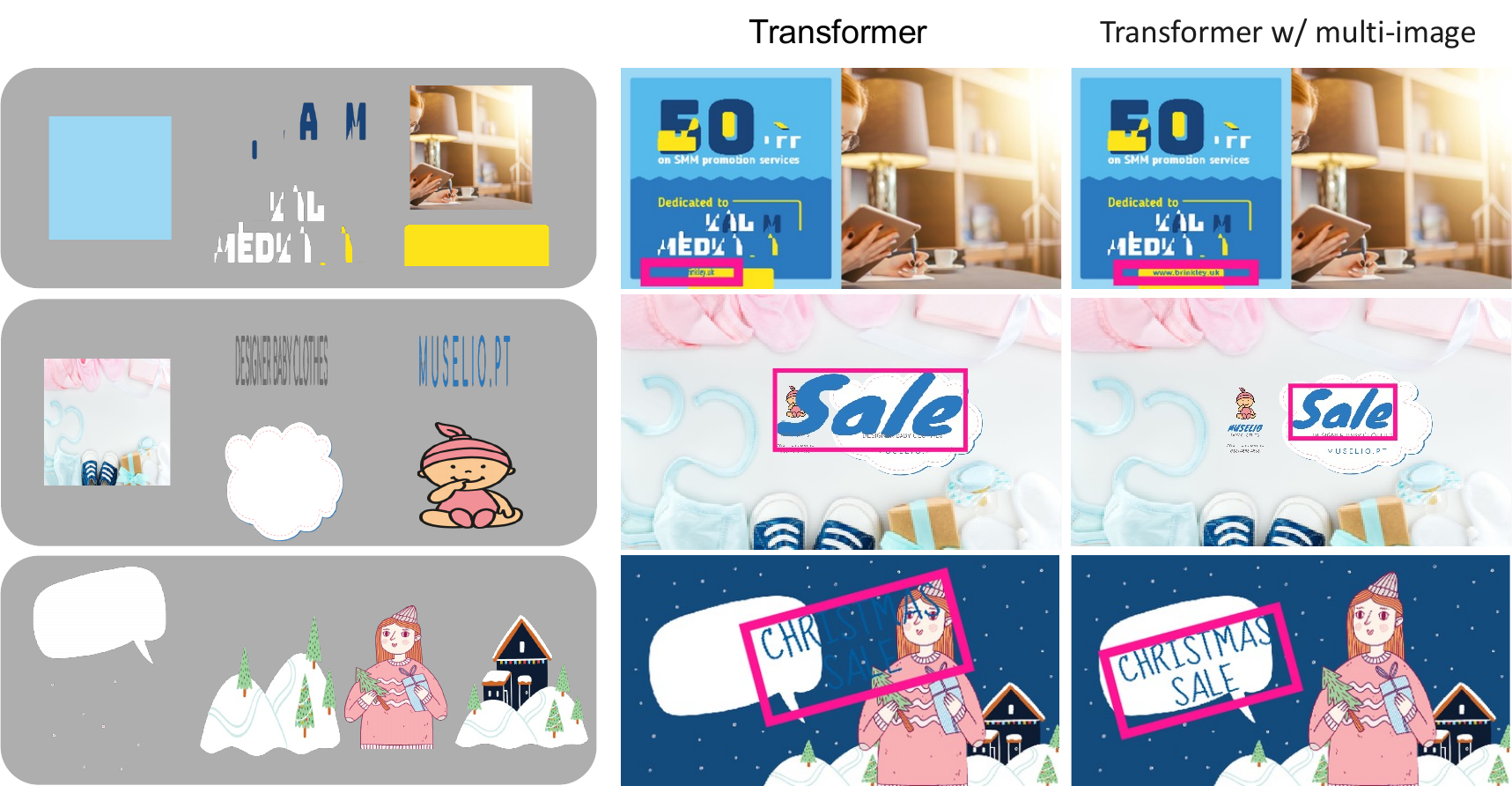}\\[-3mm]
\caption{Successful cases of Transformer with multiple image inputs. (Left) Context elements as images. (Middle) Results of Transformer with single whole image. (Right) Results of Transformer w/ multi-image.} \label{fig:TF_comparison}
\end{figure}
\section{Utilizing Multiple Image Inputs}~\label{sec:advantage_multi}



Thus far, we have explored three methods, all of which take the whole layout as a single image input. However, as shown in Fig.~\ref{fig:model-TF}(a), each element in a layout can also be rendered as an individual image (i.e., a bitmap). By providing each element’s image as an additional input, it becomes possible to explicitly learn not only the overall layout but also the visual appearance of each component and its positional relationships. This approach increases computational complexity, but the additional cost remains manageable rather than intractable through suitable encoders and linear layers.\par

We extend the standard Transformer-based method by incorporating multiple images, as shown in Fig.~\ref{fig:model-TF}(e). We refer to this approach as ``Transformer w/ multi-image.'' Each layout element, such as texts, images, and other visual components, is provided as an additional image input in this setup. By encoding these individual images, the model can more clearly capture each element's appearance and its spatial relationships within the layout. Concretely, all text and non-text elements are rendered as images and passed through a ResNet50-based encoder. The resulting embeddings, along with text attributes and layout metadata, are concatenated for each element, and then fed into the transformer to predict bounding-box coordinates, following the same procedure as the single-image transformer method.
In principle, providing multiple images as input to VLMs is also possible. However, VLMs have a maximum token limit, which practically restricts the extent to which multiple images can be utilized.
\par

Fig.~\ref{fig:TF_comparison} shows examples demonstrating the positive effect of providing multiple images as input. As shown, incorporating the appearance of each image element clarifies which elements function as ``text containers.'' For example, such containers might be uniformly colored rectangles or speech-bubble-like graphics, where text tends to be frequently placed. The Transformer model effectively learns this tendency and positions text on these containers.
\par

The bottom row of Table~\ref{tab:comparison_table} shows the performance of ``Transformer w/ multi-image,'' which achieves a substantial improvement over the version that uses only a single whole layout image. The gain is larger under the ``single text'' setting, where no other text elements besides the target text are present. This suggests that explicitly incorporating the appearance information of each image element, rather than just its coordinates or size, provides a stronger contextual cue for accurate placement. Fig.~\ref{fig:yyplot} also visualizes the performance of Transformer w/ multi-image, which achieves more near-zero BDE cases than the other methods. 
\par
Although FlexDM~\cite{inoue2023document} is capable of positioning multiple boxes, the experiment reported in its original paper evaluates only a single-box configuration under the multiple image input condition. Therefore, we present the corresponding numerical results in Table~\ref{tab:comparison_table}, which shows the standard Transformer outperforms FlexDM.
\par

\section{Conclusion, Limitation, and Future Work}
We compared three approaches for automated text box placement: a specialized Transformer-based model, a small fine-tuned VLM (Phi3.5-vision), and a large pretrained VLM (Gemini). Overall, the Transformer-based method achieved the highest accuracy, as evidenced by both IoU and BDE. Notably, BDE values often remained low across the dataset, indicating that despite occasional IoU drops -- especially with small text or complex layouts -- most placements deviated only slightly from the ground truth. This result suggests that text box placement, while challenging, is tractable with carefully tailored models.
\par

When comparing the VLM-based methods, Phi3.5-vision generally outperformed Gemini, demonstrating the advantage of task-specific fine-tuning over a purely zero-shot approach. Still, it did not match the specialized Transformer’s ability to capture spatial relationships through continuous-coordinate regression consistently. Extending the Transformer to accept multiple image inputs further lowered BDE, highlighting the benefit of incorporating richer appearance information from each layout element.
\par

For future work, several promising avenues exist. First, although we focused on a single text box, extending the proposed methods to jointly optimize multiple boxes could further enhance real-world layout design. Second, because many layouts allow more than one valid placement, representing text locations as a probability map, rather than a single deterministic coordinate, may offer a richer view of possible solutions. Third, some layouts simply do not have suitable whitespace or spacing; detecting such infeasible scenarios and suggesting alternative design adjustments would make these methods more practical. Finally, assembling multiple placement models -- e.g., specialized Transformers, VLMs, or heuristics -- into an ensemble could improve coverage and reliability, especially in challenging or unusual layout conditions.
\par

\noindent{\textbf{Acknowledgements}} 
This work was supported by JSPS KAKENHI-JP25H01149 and JST CRONOS-JPMJCS24K4.

%
%
%
\bibliographystyle{splncs04}

\bibliography{mybib}

\end{document}